\documentclass[runningheads]{llncs}
\usepackage{graphicx}
\usepackage{multicol}
\usepackage{multirow}
\usepackage{array}
\usepackage{booktabs}
\usepackage{amsmath}
\usepackage[misc]{ifsym}
\usepackage{amsfonts}

\begin{document}
\title{Vehicle Image Generation Going Well with the Surroundings}
%
%
\author{Jeesoo Kim\textsuperscript{1,$\ast$}, Jangho Kim\textsuperscript{1,$\ast$}, Jaeyoung Yoo\textsuperscript{2}, Daesik Kim\textsuperscript{2}, Nojun Kwak\textsuperscript{1}\Letter}
%
%
\institute{Graduate School of Convergence Science and Technology\\ Seoul National University, Suwon 16229, Republic of Korea\\
\email{\{kimjiss0305,kjh91,nojunk\}@snu.ac.kr}\\
\and
Naver Webtoon, Republic of Korea\\\email{\{yoojy31,daesik.kim\}@webtoonscorp.com}}

\maketitle              
\begin{abstract}
In spite of the advancement of generative models, there have been few studies generating objects in uncontrolled real-world environments.
In this paper, we propose an approach for vehicle image generation in real-world scenes.
Using a subnetwork based on a precedent work of image completion, our model makes the shape of an object.
Details of objects are trained by additional colorization and refinement subnetworks, resulting in a better quality of generated objects.
Unlike many other works, our method does not require any segmentation layout but still makes a plausible vehicle in an image.
We evaluate our method by using images from Berkeley Deep Drive (BDD) and Cityscape datasets, which are widely used for object detection and image segmentation problems.
The adequacy of the generated images by the proposed method has also been evaluated using a widely utilized object detection algorithm and the FID score.

\end{abstract}

\footnotetext[1]{Equally contributed}

\section{Introduction}

Most of the recent advances of object generation models \cite{ref_article1,ref_article6} often implicitly assume some rules when learning a domain such as aligned components and fixed location of an object.
To compose an image and arrange objects in a non-trivial location, several works have used the semantic layout to transform it into an image of a real scene.
One of the frequently used datasets in this task is \textit{Cityscape} \cite{ref_article3} which includes road scene images with pixel-wise annotations.
Many researchers have made remarkable progress and succeeded to generate a realistic synthetic image using a semantic layout~\cite{ref_article10,ref_article2,ref_article13}.
Using these techniques, Hong \textit{et al.}\cite{ref_article8} has proposed a method that can generate an object in an arbitrary location assigned by a user using a semantic layout.
However, training a model and synthesizing an image using the semantic layout is an inefficient method since making a pixel-wise annotation requires a lot of time and human effort.
Demand for an inexpensive annotation in computer vision tasks is being raised steadily and one example is the weakly supervised object localization which tries to localize images only using image-level labels \cite{ref_article17}.

Meanwhile, image completion is the task of filling a missing area in an image.
Research using convolutional neural networks(CNN) has made a breakthrough in this field by making the models learn features around the deleted area \cite{ref_article9,ref_article12}.
Based on the clues given around the deleted area such as the corners of the table or the bottom side of the window of a house, models are able to complete the image naturally.
However, with no other condition given, generating an object is impossible if no clue is given outside the deleted area.
Also, they usually fail to complete some complicated texture such as the face of a human or a portion of animals.
Training these models with a single object category, they are capable of making the coarse outline of the object but lack details when not enough clues from its surroundings are given.
Since the appearance of the objects in image inpainting problems is not consistent like the images of \textit{MNIST} or \textit{Celeb}, image inpainting methods highly rely on the reconstruction loss rather than the adversarial loss which makes the object sharper.
For this reason, the existing inpainting methods are very hard to generate a sharp image with an object inside.

In this paper, we propose a method to generate vehicles in given locations, which understands typical appearances of vehicles but with various orientations and colors.
With a box given at an appropriate location with an appropriate size, our model can generate learned objects inside.
The generated cars suit the surrounding background and even complete occluding objects ahead.
Note that all we need to generate an object is just a box annotation while many other works in image translation require a segmented target map to transform it into a realistic image.
Our proposed method simply needs a rectangular-shaped mask to generate a vehicle, regardless of the location it is placed.
For this purpose, we utilize Berkeley Deep Drive (BDD) dataset \cite{ref_article14} which contains box annotations around the objects in the images.
It is widely used for the tasks of object detection and image segmentation. 
By training the car areas annotated in the BDD dataset, we can generate a vehicle that goes well with its surroundings in an arbitrary location.
The contributions of this paper are as the followings:

\begin{itemize}
\item We propose a method that generates a vehicle object in an image that goes well with the background.
\item Unlike other methods using pixel-wise annotated information, only box annotations are used as a given knowledge.
\item Our method is capable of generating vehicles over an empty road image as well as substituting the existing vehicles.
\end{itemize}

\section{Related works}
In order to synthesize a lifelike image, many researchers utilized label maps or contours as a guide to generate a real image.
Isola \textit{et al.}\cite{ref_article10} proposed a method to transform any kind of blueprint into the desired output using GAN.
Chen \textit{et al.}\cite{ref_article2} adopted a feature learning method by defining layer-wise losses.
This enabled transforming a high-resolution semantic layout into an actual image.
Patch-based methods such as the work of Wang \textit{et al.} \cite{ref_article13} enabled high-quality image translation with an image size more than 512 $\times$ 512.
Generating objects in locations decided by a user has been possible in the work of Hong \textit{et al.} \cite{ref_article8} using generative adversarial network.
However, input layouts used in all of these methods are expensive to prepare in real-world since they should be annotated pixel-wise,
which requires much more cost than just annotating bounding boxes.

Studies trying to generate more complicated context using clues in the image have been proposed recently.
Image completion is the task of filling an appropriate image segment into a vacated region of a real-world image.
The model must learn to paint the image through the blank area.
For example, in an image of a building, the corner of a window is removed and then the model must fill it up with glass and bricks aside.
Coming to an image of a human face, one of the eyes or the nose is erased and the network recovers it in a natural form.

The Context Encoder (CE), an early work of image completion using a neural network \cite{ref_article12}, has been proposed for this problem.
Using a channel-wise fully connected layer between an encoder and a decoder, CE roughly generates pixels in the image using the encoder features.
The work of Iizuka \textit{et al.}\cite{ref_article9} has approached this problem using dilated convolutions and a specific type of discriminator.
The dilated convolution acts just the same as the original convolution except that the filter is applied to the feature skipping a few pixels or more.
This derives the expansion of the receptive field without loss of resolution.
Although both approaches can recover the missing parts of an image, it is impossible to generate a whole object.
Even if trained only with a particular object, the result still suffers from poor performance.
This is because the model bumbles among many choices of how the object will be generated.
Since the completion model can predict the shape and surrounding context roughly, we use it as a primary module of our method.

\section{Approach}
In this paper, we assume that every object can be expressed using two traits, shape and texture.
Our method consists of two consecutive networks learning each of these properties.

\begin{figure*}[t]
	\centering
	\includegraphics[width=0.8\linewidth]{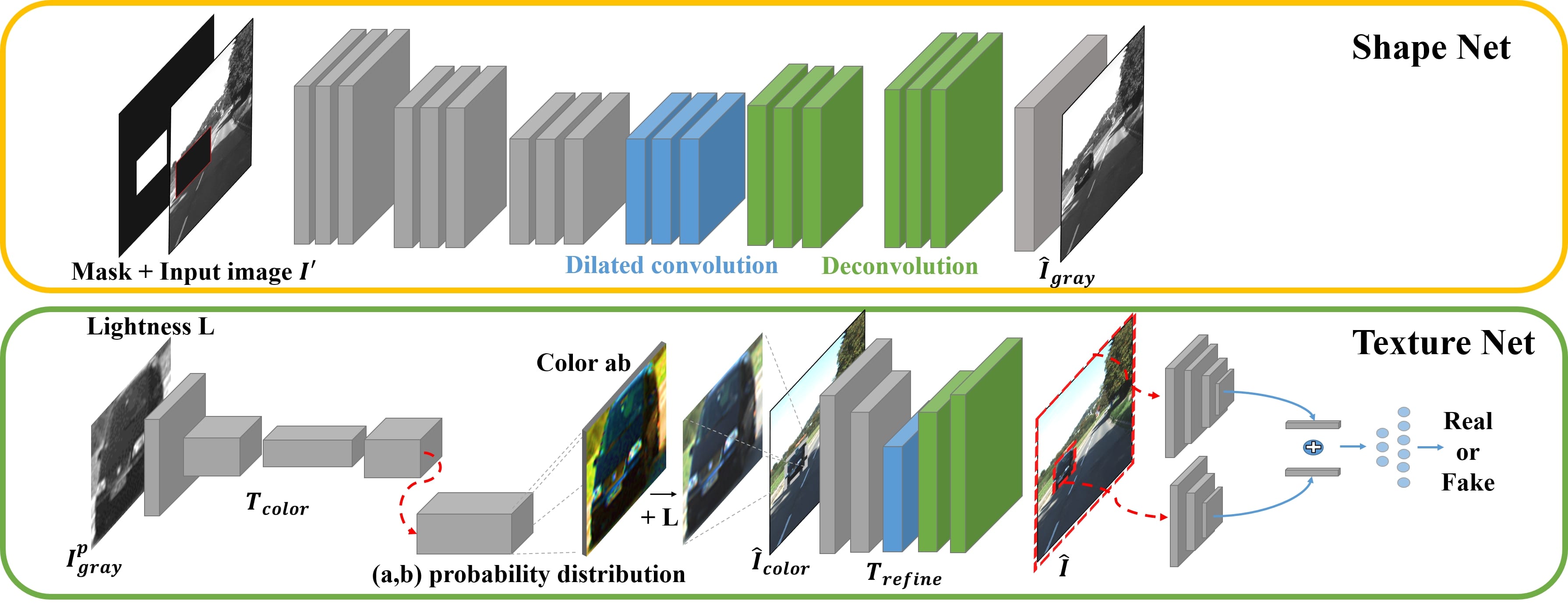}
	\caption{Overall architecture. The top-side of the figure shows the Snet and the bottom-side shows the Tnet which is composed of $T_{color}$ and $T_{refine}$. Transposed convolutional layers are used to recover the resolution of $\hat{I}_{color}$.}
	\label{fig:arch_total}
\end{figure*}

\subsection{ShapeNet}

As we assume that the segmentation layout is not given, we adopt an image completion model which understands the surrounding texture around the erased area.
The architecture used in our ShapeNet (Snet) is the same as the work of Iizuka \textit{et al.}\cite{ref_article9}.
Snet maintains the feature resolution at a certain level rather than going down to a vector-level representation, which is just a quarter of the original image size in our case.
For the image completion task, the information from the features nearby is more important.
Therefore, decreasing the resolution only to half is adequate to fill the blank area.
Also, to extend the receptive field of the features in each layer, dilated convolution \cite{ref_article15} is used in the middle of the network.
In this way, the network can reflect the information not only from the pixels adjacent to the blank area but also from those far away.
This highly encourages the network to decide how the object should be placed and how the occluding object should be completed.
At the end of the Snet, the network reconstructs the gray-scaled image of the original input so that the model concentrates on constructing the pose and shape of the generated object, not the color or details of the context.
The detailed procedure of Snet can be summarized as follows:
\begin{equation}
  \hat{I}_{gray} = S([I'; M]).
  \label{equ:snet}
\end{equation}
Here, $I\rq$ represents the image of which an area corresponding to the mask $M$ has been removed and $S(\cdot)$ denotes the Snet.
Note that the mask $M$ is not a segmentation layout but a rectangular binary map which corresponds to the bounding box of the object.
The encoding step includes 2 steps of downsampling and 4 types of dilated convolution (rate = 1, 2, 4, 8).
The decoding step upsamples the features into the original scale using transposed convolutional layers.
Pixels outside the generated region are preserved as we do not wish any change outside of the box.
After the Snet generates an image, only the region inside the box is combined with the stored background pixels.
By doing this, only the change that occurred inside of the box is taken into consideration.
The Snet is trained to minimize the $L_1$ distance 
between the synthesized grayscale image $\hat{I}_{gray}$ and the original gray scale image $I_{gray}$ as in (\ref{equ:snetobj}).
The Snet is trained individually beforehand.
\begin{equation}
  L_{S} = \|I_{gray}-\hat{I}_{gray}\|_{1}.
  \label{equ:snetobj}
\end{equation}

\subsection{TextureNet}
\label{sec:tnet}
Complicated visual properties of real objects often disrupt the learning of generative models.
In the problem settings of this paper, the overall structure and the detailed texture including colors can be difficult to depict by a single generator.
When it comes to generating a car, details such as headlights, taillights, or the windshield of a car may vary a lot. 
If the Snet makes a rough sketch, the TextureNet (Tnet) adds details by painting colors on the sketched grayscale image and refines it.
We have adopted the architecture in \cite{ref_article16} which was used for a colorization task.
The Tnet can be summarized as follows:
\begin{equation}
  \hat{I}_{color} = T_{color}(I_{gray}^{p}, I')
  \label{equ:tnetcolor}
\end{equation}
\begin{equation}
  \hat{I} = T_{refine}(\hat{I}_{color})
  \label{equ:tnetrefine}
\end{equation}
\begin{equation}
  L_{T} = CrossEntropy(I, \hat{I}_{color}) + \lambda\|I-\hat{I}\|_{1}.
  \label{equ:tnetrefineobj}
\end{equation}

Tnet is divided into two components: $T_{color}(\cdot)$ which paints colors on the gray patch image $\hat{I}_{gray}^{p}$ and paste it back to the masked image $I'$, and $T_{refine}(\cdot)$ which corrects the details of the colorized image $\hat{I}_{color}$ as in (\ref{equ:tnetcolor}) and (\ref{equ:tnetrefine}), respectively.
After the $T_{color}$ classifies the color classes of all pixels, the colorized image of $\hat{I}_{gray}$ is pasted back to $I$, which produces $\hat{I}_{color}$.
We crop the patch image $I_{gray}^{p} = Crop(I_{gray},M)$ where the object is to be generated since the remaining region does not need any colorization nor manipulation.
In the Tnet, $T_{color}$ in (\ref{equ:tnetcolor}) learns the color distribution of cars in detail.
As in \cite{ref_article16}, $T_{color}$ predicts the values of each pixel in the CIELab color space.
Since the grayscale image corresponds to the lightness $L$, the model should predict the remaining a and b values.
A pretrained VGG-19 network is used to extract features from the gray image.
After that, a 1$\times$1 convolution filter outputs a tensor with a depth of 313.
Dividing the ab color space into 313 bins, the $T_{color}$ turns the colorization problem into a classification problem which is much easier than directly generating images having various occlusions and a wide color variation.
The cost is calculated by the cross-entropy function between the output and the Lab-encoded ground-truth values as in the first term of (\ref{equ:tnetrefineobj}). 
The color is painted solely on the patch image, which can be awkward when the patch is pasted back to the original image.
Therefore, we use an additional network module $T_{refine}$ that encourages the object to get along with the surroundings of the image.
Using the same structure with the Snet highly sharpens the blurry results of colorized images and adaptively optimizes the colors of the neighboring objects such as occluding cars, road segments or parts of buildings.
This module is trained adversarially using a global-local discriminator along with a reconstruction L1-loss as shown by the second term in (\ref{equ:tnetrefineobj}).
The value of $\lambda$ is set to be 0.5 in our work.
The colorization network offers important clues for the refinement network such as the location of taillights and the rear window, helping the refinement network concentrate on polishing the details.

\begin{figure*}[t]
	\centering
	\includegraphics[width=0.7\linewidth]{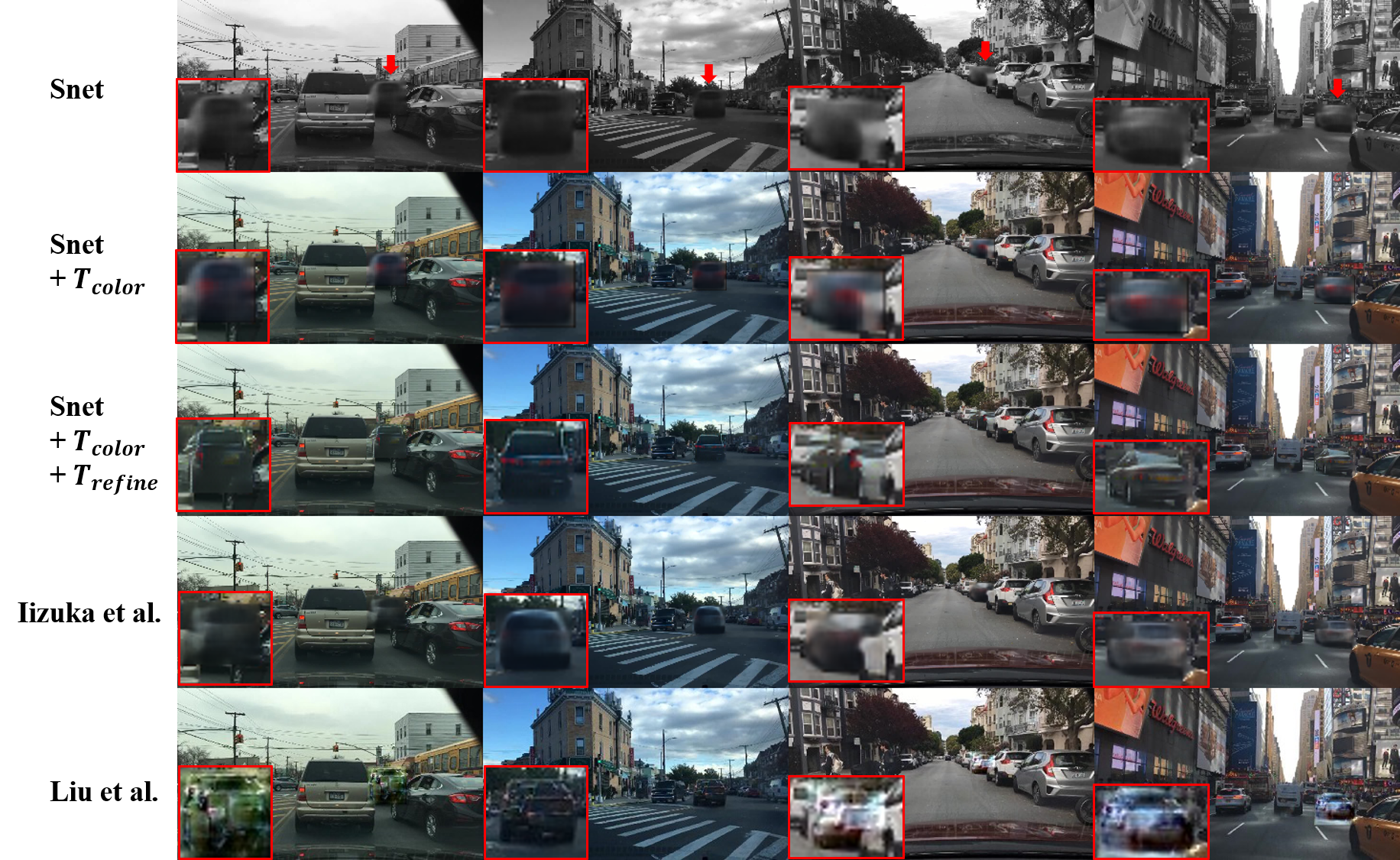}
	\caption{Results from our method and the baseline \cite{ref_article9}. From top to bottom, 1) completed gray-scaled results of Snet, 2) results colorized by $T_{color}$, 3) final results of Tnet after refinement, and 4, 5) results of the baselines. Compared to the results of the baselines, our model depicts the details better since the refine module of Tnet can produce elaborate outputs. To present how well our method performs by itself, no post-processing is applied to all results.}
	\label{fig:ablation}
\end{figure*}

\subsection{Global-Local discriminator}
\label{sec:discriminator}
A global-local context discriminator is used to train our model in the same way as in \cite{ref_article9}.
The overall discriminator $D$ consists of one global discriminator $D_{global}$ and one local discriminator $D_{local}$.
The global discriminator takes the entire image $I$ (real) or $\hat{I}$ (generated) as an input and the local discriminator takes the patch image $I_p$ or $\hat{I}_p$ inside the annotated box from the respective image $I$ or $\hat{I}$. The size of the patch image is normalized to 64$\times$64 by ROI-pooling.
Each discriminator is followed by a fully connected layer.
After that, both features are concatenated into a single vector and then processed by a single fully connected layer.
A sigmoid output produces an outcome which is trained to correctly decide whether the given image is real or completed by the model.
This encourages the model to depict sophisticated details of the object. The discriminator $D$ and $T_{refine}$ are trained in an adversarial way as follows:

\begin{equation}
\label{equ:gldis}
\begin{split}
\min_{T_{refine}} \max_{D} & \quad  (\mathbb{E}_{I_{p},I\sim P_{data} } [\log(D(I_{p},I))] \\
 &+ \mathbb{E}_{\hat{I}_{color}\sim P_{T_{color}(\hat{I}_{gray})}} [1-\log(D(\hat{I}_p, \hat{I}))])
\end{split}
\end{equation}

We may consider the colorized image input $\hat{I}_{color}$ from the $T_{color}$ as a noise.
This allows our model to be trained adversarially.

\begin{figure*}[t]
	\centering
	\includegraphics[width=0.7\linewidth]{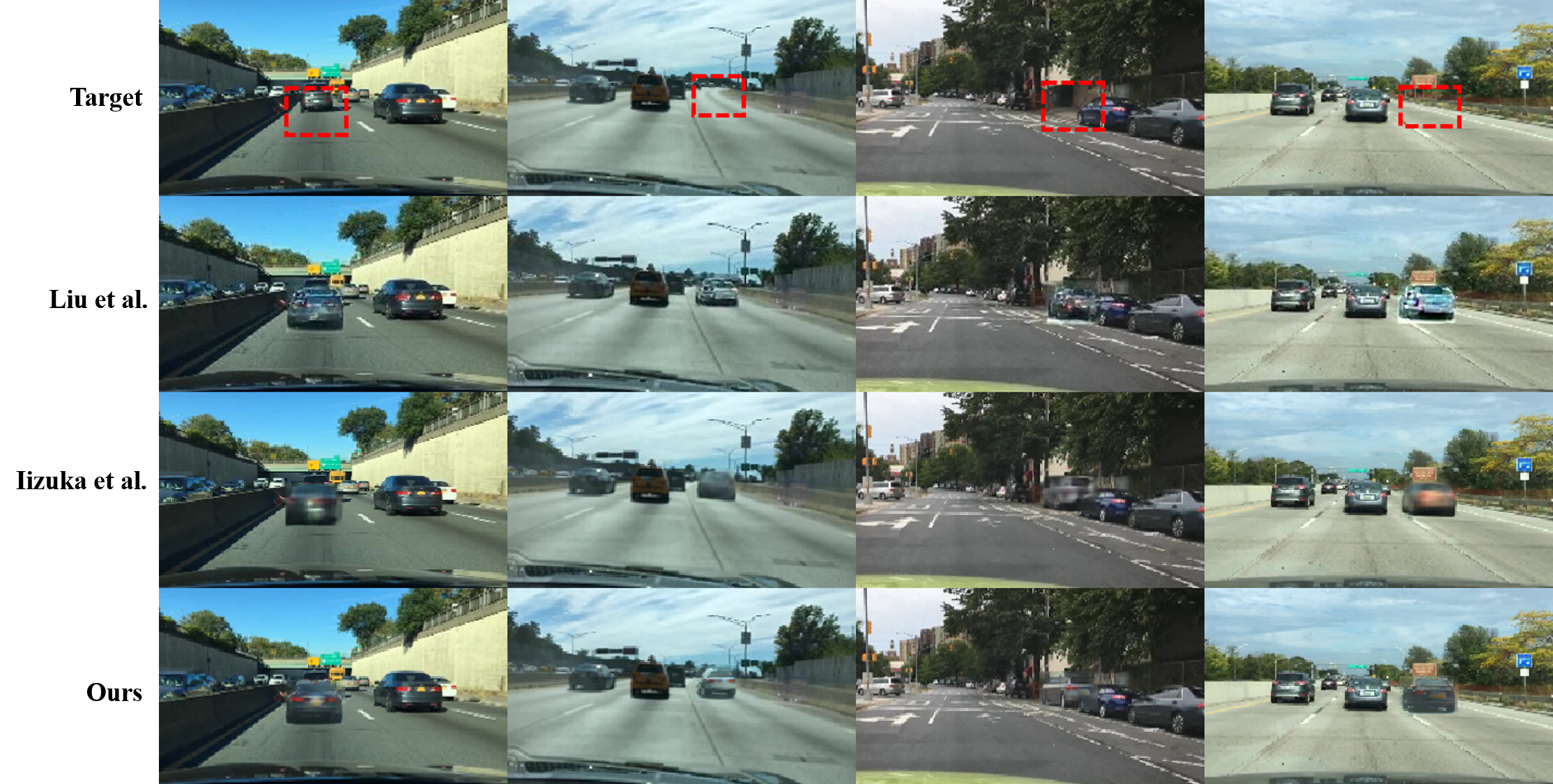}
	\caption{Generating vehicles on empty roads. Target regions where the vehicles are to be generated are marked by red rectangles in the first row. With proper boxes given, both the baselines and our method draw vehicles inside. 
	}
	\label{fig:emptyroad}
\end{figure*}

\begin{figure*}[t]
	\centering
	\includegraphics[width=0.7\linewidth]{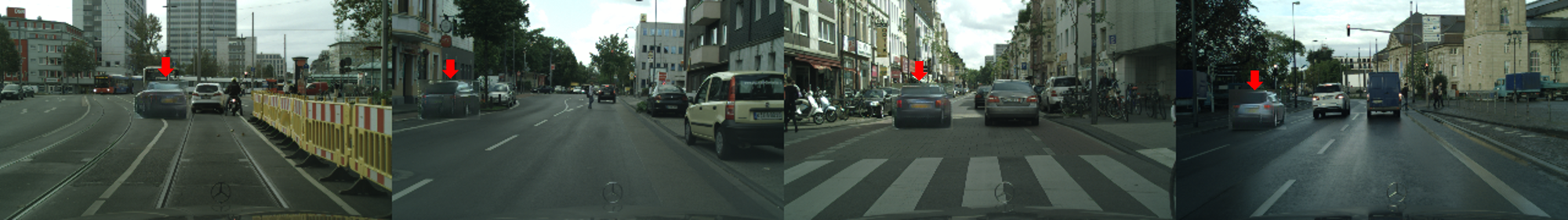}
	\caption{Our results generated over the \textit{Cityscape} dataset. The red arrows indicate where the vehicles are generated. }
	\label{fig:cityscape}
\end{figure*}

\section{Experiments}

\subsection{Vehicle generation subject to the surroundings}
Among various objects, we have chosen to specifically concentrate on the car which has more clear visibility compared to other objects in the data.
BDD100K is a dataset including videos filmed by running cars \cite{ref_article14}.
After resizing the resolution to $320 \times 180$ for the stable training, boxes sized under 10 pixels either horizontally or vertically are excluded in the training since they are so small that it is hard to acknowledge them as cars, which makes it meaningless to generate them.
Also, a large vacated area is known to be hard to fill by the image completion frameworks.
Therefore, we exclude boxes bigger than $64 \times 50$ since both the baseline and our method show poor results on them.
We evaluated our trained model by generating vehicles on test images of BDD100K dataset.
Since there is no other research generating an object in a given position only using the box annotation, we set the research of \cite{ref_article9} and \cite{ref_article11} used in the image completion problem as the baseline models.
As the image completion problem is quite similar to our problem, 
the baseline holds the capacity to generate objects if trained properly.
Instead of vacating a random region of the image, we only deleted the region where a vehicle exists and regenerated a new vehicle.
As the baseline models only witness particular appearances, this makes them possible to generate an object, though they fail to express the details of objects as explained in this paper.

Some results of \cite{ref_article9} are shown in the fourth row of Fig.~\ref{fig:ablation}.
Cars generated at the center of the images mostly show their backside since most of the cars are running forward.
For the boxes located between the road and the sidewalk, the cars are generated slightly askew as the camera usually shoots them in the diagonal direction.
The overall appearances of all results resemble the real vehicles but the details such as taillights and wheels lack delicacy.
Despite using the global-local discriminator described in Section \ref{sec:discriminator}, the baseline model generates blurry results.
Excessive abuse of adversarial loss easily corrupts the model and impedes the training, which is why we mitigated the problem by our method.
Results from the work of \cite{ref_article11}, the last row in Fig.~\ref{fig:ablation}, are considerably poor compared to other methods.
Using the perceptual loss introduced by Gatys \textit{et al.} \cite{ref_article5}, the model produces reasonable background segments and makes an appropriate outline of a vehicle.
However, the texture inside is quite noisy, which seems to draw cars inside a car, which are hard to be perceived as cars.
The third row of Fig.~\ref{fig:ablation} shows some results of our method.
In our method, the outline of vehicles is clearly visible and parts such as the taillights and wheels are described delicately.
Also, the texture around the generated car suits the surrounding background.
The paved roads are naturally connected at the bottom side with shadows shaded by the body of the vehicle.
Furthermore, the model can recognize a building or a bush behind and reflect it in the background.
Especially, for the cars tailing back aside from the road, the generation of an occluding car ahead is possible (see the third column in Fig.\ref{fig:ablation}).

Although not presented in this paper, we also have tried to use the refine network module ($T_{refine}$) directly after the Snet producing colorized images instead of black and white images.
However, the adversarial loss easily corrupts the model making the images unrecognizable.
In Fig.\ref{fig:ablation}, the results of Snet (the first row) and Iizuka \textit{et al.} (the fourth row) are similarly blurry and incomplete.
Adding clues of texture by colorization (the second row) highly helps the training of $T_{refine}$ in which the adversarial loss is dominant.
Also, while $T_{color}$ focuses on painting the vehicle regardless of its surroundings, $T_{refine}$ makes up for the inconsistent surroundings.

\subsection{Vehicle generation on an empty road}
After training the model, we experimented with the situation in which our model generates cars at given locations where no car existed in the original images, which is the eventual objective of this paper.
Images of the road with no cars on it are chosen, which would have been never used in training since it has no annotation box on it.
Fig. \ref{fig:emptyroad} shows the results of how well our method generates objects on empty roads where we asked it to.
Given a reasonable location, the model is capable of generating a car with a random context.
Especially, our method shows a remarkable performance when it comes to generating a car parking alongside the road.

\begin{figure}[t]
	\centering
	\includegraphics[width=0.7\linewidth]{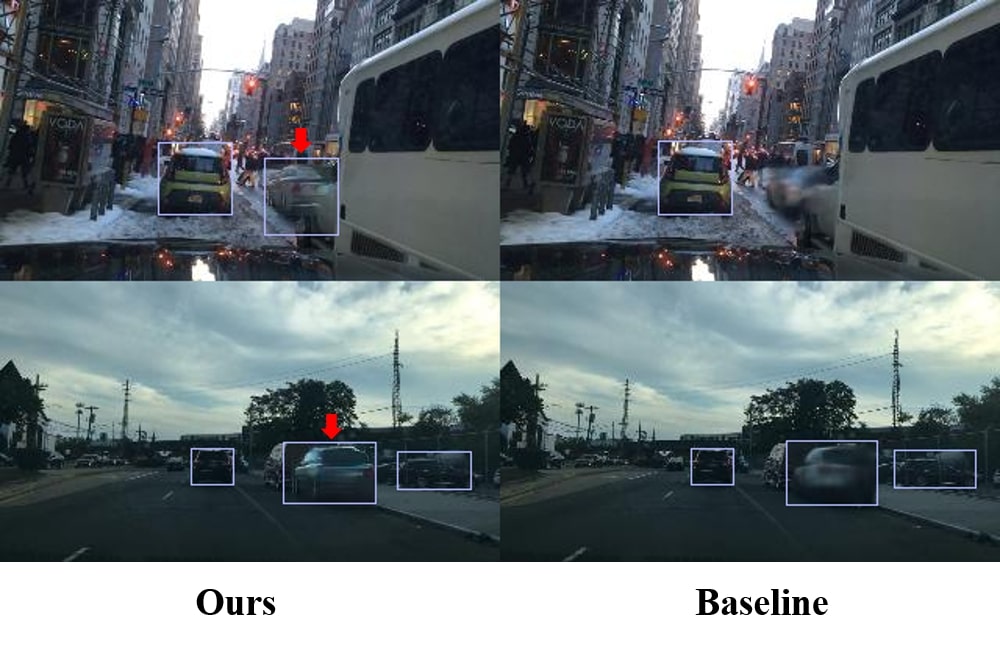}
	\caption{Detection results of generated objects. Boxes in purple show the detection result of SSD. The top row shows the generated sample that our method succeeded to deceive the detector to perceive it as a vehicle while the baseline failed and the second row shows the case where both methods deceived the detector in spite of the poor quality of the baseline. In spite of our far better quality, SSD occasionally detects the deformed results of the baseline as a car.}
	\label{fig:detection}
\end{figure}

\begin{table}
  \centering
  \caption{Object detection score (recall in \%) when Tnet is additionally used (Ours) or not (Baseline). Single Shot multibox Detector(SSD) \cite{ref_article18} is used to evaluate how well the objects are generated. Weights of SSD are pre-trained using PascalVOC dataset. We only report the recall score at the target region since vehicles outside are not under consideration.}
    \begin{tabular}{@{}ccc@{}}
    \toprule
    \multirow{2}{*}{Method} & \multicolumn{2}{c}{Confidence threshold} \\ \cmidrule(l){2-3} 
                            & 0.12                & 0.3                \\ \midrule
    Original BDD            & 87.42               & 78.59              \\
    Baseline(Iizuka \textit{et al.})                & 63.35               & 50.23              \\
    \textbf{Ours}           & \textbf{73.12}      & \textbf{60.24}     \\ \bottomrule
    \end{tabular}
\label{tab:ssd}
\end{table}
\subsection{Vehicle generation on Cityscape dataset}
To show that our method is not restricted to the dataset it is trained on, we apply our network trained by BDD to generate cars on images from the \textit{Cityscape} dataset \cite{ref_article3}.
Since, both of the datasets contain road scene images in the view of the drivers, the perspective and overall aspects are quite similar except that BDD is taken in the U.S.A and \textit{Cityscape} in Germany.
Fig. \ref{fig:cityscape} shows the results of our model applied to the \textit{Cityscape} dataset.
Without the red arrow indicating the generated vehicle, it is difficult to recognize which one is the generated vehicle at one glance.

\begin{table}
\caption{The FID scores of the methods used in this paper. Only the generated region is used for the evaluation since the rest of the area is not synthesized.}
\centering
\begin{tabular}{cclll}
\cline{1-2}
Method                                      & \multicolumn{1}{l}{FID score} \\ \cline{1-2}
\multicolumn{1}{l|}{Iizuka \textit{et al.}}    & 112.3                           \\
\multicolumn{1}{l|}{Liu \textit{et al.}} & 50.54                         \\
\multicolumn{1}{l|}{\textbf{Ours}}             & 41.04                         \\ \cline{1-2}
\end{tabular}
\label{tab:fid}
\end{table}

\subsection{Effectiveness of using Texture net}
Bringing the global-local discriminator to the end of Snet and changing the output dimension to RGB channels is equivalent to the work of \cite{ref_article9}.
Though we have shown the qualitative improvement in the figures above, we report an additional numerical comparison to prove the effectiveness of Tnet.
For a fair evaluation of how much our TextureNet contributes to the training, we apply the Single Shot multibox Detector (SSD) \cite{ref_article18}, a widely used object detection algorithm, to the images generated by our method and the baseline\cite{ref_article9}.
Vehicles from the original images are substituted by both methods.
If objects are generated properly, the detector should be able to locate them in the image, meaning that the generated objects are assumed to be sampled from the data distribution it is trained.

Our model and the baseline \cite{ref_article9} have been trained using the training data in BDD100K and the evaluation has been carried out with the validation data.
Since only the object to be generated is under consideration, vehicles detected outside the generated region are ignored. 
Therefore, we only evaluate the recall of SSD, which is the ratio of whether the detector finds the generated object or not.
In the dataset, there are a total of 6,020 vehicles to recall that satisfy our size constraints.
Since the detector is trained using PascalVOC dataset \cite{ref_article4}, it is fair enough to compare the baseline and our method applied to BDD100K dataset.

Table \ref{tab:ssd} shows the recall scores of SSD on BDD dataset.
For the original BDD images, SSD detects vehicles in the box area at a rate of 78.59\% using the confidence threshold of 0.3 and 87.42\% using the confidence threshold of 0.12 as shown in Table \ref{tab:ssd}.
In both class confidence score thresholds, our method highly precedes the baseline by about 10\% points.
Additionally, we analyze the detection result of which the threshold is 0.12.
Samples, which our method successfully makes the detector perceive as vehicles while the baseline fails to, occupy 17.24\% while the opposite case records 7.47\%.
One example of this case is shown in the top row of Fig.~\ref{fig:detection}.
Meanwhile, though both methods successfully deceive the detector in many cases, quite a lot of samples generated by the baseline suffer from a qualitatively poor performance.
This case is presented in the bottom row of Fig.~\ref{fig:detection}.
Considering that some ambiguous results from the baseline are occasionally detected by the detector, we expect that the performance of our method would be rated higher if evaluated by a qualitative metric.

\subsection{Measuring generation quality}
Frechet inception distance (FID) score \cite{ref_article7} is a measure frequently used to evaluate the generation quality of generative models.
Using a pre-trained Inception-V3, FID compares the statistics of generated images to the real ones.

In Table \ref{tab:fid}, we assess the quality of images generated by each method.
Images from \cite{ref_article9} roughly have the appearance of vehicles but record the worst FID score.
This attributes to the abstractly depicted context of each vehicle.
The work of Liu \textit{et al.} shows an intermediate level of FID score.
Though the silhouette of the generated sample resembles the vehicle, the texture inside is rather a noise which causes an increase in the FID score.
Our method has the lowest FID score, which means that our method generates samples that have the most similar distribution to that of real samples.

\section{Discussion}
\begin{figure*}[t]
	\centering
	\includegraphics[width=0.7\linewidth]{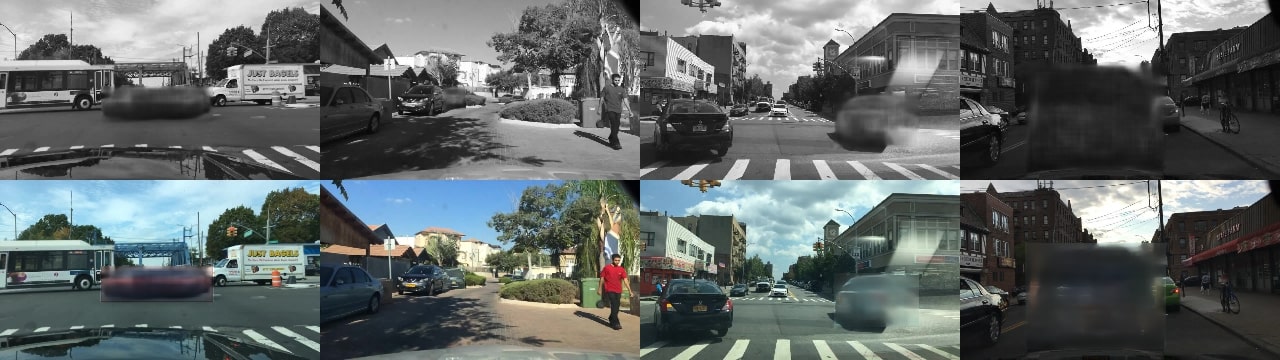}
	\caption{Failure cases. On the first row lie some single channel outputs of Snet and on the second row lie the corresponding 3-channel outputs of Tnet.}
	\label{fig:failure}
\end{figure*}

The object detection dataset includes annotations indicating the location where the car exists, not where the car \textit{may} be.
This enforces us to give the model a box at a reasonable position manually.
Although the most important merit of our method is that we only use a box annotation to generate a vehicle, the necessity of giving a box candidate still remains.
For example, with a box given upon a building, the model would still try to generate a car even if it is nonsense.
With a highly predictive scene understanding model that can suggest the location where the vehicle may be, we expect a fully automated vehicle image augmentation would be possible. 
Also, wide boxes, which correspond to the cars near the camera, usually fail to be filled, which is the drawback of the image completion models.
A portion of the car may lie outside the camera angle and this will go against the overall aspect of trained objects.
Some of the failure examples are given in Fig.~\ref{fig:failure}.

\section{Conclusion}
We tackled a problem to generate vehicles at designated locations over images of real scenes.
To solve this problem, we used an architecture composed of two subnetworks which generate the shape and the texture of the vehicle respectively. 
The Snet roughly completes the shape of the car according to the surroundings while the Tnet decides the details of the generated vehicle resulting in a better generating performance.
Consequently, our method can generate objects going well with the surroundings in arbitrary locations.
This is highly expected to be foundation research for image augmentation research.

\section{Acknowledgements}
This work was supported by the National Research Foundation of Korea (NRF) grant funded by the Korea government (2021R1A2C3006659).


\end{document}